\documentclass[12pt,authoryear]{elsarticle}

\usepackage{graphicx}
\usepackage{epstopdf}
\usepackage{url}
\usepackage[onehalfspacing]{setspace}

\usepackage{soul}
\usepackage{color}

\newcommand{\edits}[1]{#1}
\newcommand{\replace}[2]{#2}

\journal{Journal of Biomedical Semantics}

\begin{document}
	
\begin{frontmatter}
	
\title{\replace{Higher order features}{Word embeddings} and recurrent neural networks based on Long-Short Term Memory nodes in supervised biomedical word sense disambiguation}

\author[label1]{Antonio Jimeno Yepes}
\address[label1]{IBM Research Australia, Melbourne, VIC, Australia}

\begin{abstract}

Word sense disambiguation helps identifying the proper sense of ambiguous words in text.
With large terminologies such as the UMLS Metathesaurus ambiguities appear and highly effective disambiguation methods are required.
Supervised learning algorithm methods are used as one of the approaches to perform disambiguation.
Features extracted from the context of an ambiguous word are used to identify the proper sense of such a word.
The type of features have an impact on machine learning methods, thus affect disambiguation performance.
In this work, we have evaluated several types of features derived from the context of the ambiguous word and we have explored as well more global features derived from MEDLINE using word embeddings.
Results show that word embeddings improve the performance of more traditional features and allow as well using recurrent neural network~\edits{ classifier}s based on Long-Short Term Memory (LSTM) nodes\replace{, which further improve the disambiguation performance}{}.
The combination of unigrams and word embeddings \edits{with an SVM} set\edits{s} a new state of the art performance with a\replace{n}{macro }accuracy of 95.97 in the MSH WSD data set.

\end{abstract}

\begin{keyword}
Word sense disambiguation \sep Word embeddings \sep Recurrent neural networks \sep Biomedical domain
\end{keyword}

\end{frontmatter}

\section{Introduction}

The amount of biomedical text published is growing exponentially and researchers are finding it increasingly difficult to find relevant information.
The automatic processing of biomedical articles can help with this problem by identifying biomedical entities (such as genes, diseases, drugs), and the relations between them.
This information can be extracted from text and used for applications such as summarization, data mining and intelligent search.
However, identifying biomedical entities and relations in text is a complex and challenging task.

One difficulty, addressed by this research, is the problem of lexical ambiguity.
Lexical ambiguity is the presence of two or more possible meanings within a single term or phrase. For example, determining whether the term \textit{bass} is referring to a~\textit{fish} or~\textit{instrument} given the context in which the term is used.
Disambiguation is useful in concept mapping algorithms and tools relying on dictionary look up, such 
as MetaMap~\citep{aronson2010overview}.

The goal of word sense disambiguation (WSD) is to automatically predict the most likely sense of an ambiguous word.
\edits{For instance, the word~\textit{cold} could refer to the temperature or an infection depending on the context in which it is used.
A WSD algorithm would predict the most appropriate sense given the context of the ambiguous word.}

There are several approaches being used for WSD which range from supervised approaches (which rely on examples of use of each ambiguous word in context to train a learning algorithm) to knowledge-engineering approaches (which rely on a sense catalogue such as a dictionary).

In this work, we explore the use of word embeddings as candidate representations for the WSD problem.
We show that unigram representation is a strong baseline using Support Vector Machines as the machine learning algorithm, but that neural network word embeddings improve theses baseline results.
We explore as well the different parameters used in the generation of word embeddings.

\replace{In addition, results are significantly improved when using word embeddings with recurrent neural networks as the learning algorithm.}{}
Furthermore, a combination of word embeddings and unigam features with SVM set a new state of the art disambiguation in \edits{macro }accuracy of 95.97 in the MSH WSD data set.

\section{Related Work}

WSD algorithms utilize the context in which a term is used to identify the appropriate sense of a lexical ambiguity.
Existing disambiguation algorithms to resolve ambiguity can be divided into three groups: unsupervised~\citep{Pedersen10,Brody09,Chasin14}, supervised~\citep{Zhong10,Stevenson08}, and knowledge-based~\citep{Navigli11,Agirre10,McInnes11,Jimeno11} algorithms.
Unsupervised algorithms typically use clustering techniques to divide 
occurrences of an ambiguous word into groups that are later associated 
with their possible sense and might help identify new senses~\citep{lau2012word}.
Supervised algorithms use machine learning techniques to assign concepts 
to instances containing the ambiguous word, thus these methods require examples of use of the different senses of the ambiguous words for model training.
Knowledge-based algorithms do not require a corpus containing examples of the 
ambiguity but rather use information from an external knowledge source such 
as a taxonomy or dictionary.

In this work, we focus on supervised learning algorithms with the intention of exploring higher order features. 
Even though developing data sets for supervised methods is expensive, we believe that the insights learned by exploring features with supervised methods can be beneficial for other kinds of methods.



As in many supervised learning tasks, representation of the problem is relevant to the performance of the task~\citep{jimeno2015feature}, i.e. transforming text into features to be used by machine learning algorithms.
There are several feature sets used in previous work~\citep{agirre2007word, navigli2009word}, this includes local features, topical features and syntactic dependencies.

\citep{Stevenson08} have shown that using linguistic features in combination with meta-data of the published articles (e.g. MeSH{\tiny $^{\textregistered}$} headings) improve disambiguation performance, even though manually annotated meta-data features cannot be assumed to be always available.
\citep{mcinnes2007using} used the annotation provided by MetaMap to automatically assign UMLS{\tiny $^{\textregistered}$} concept identifiers. 
Overall, using additional features to unigrams improves the WSD performance.

The features engineered in previous work on biomedical WSD have focused on local features derived from the context of the ambiguous word or meta-data of the citation. 
We would like to take a step further and consider higher order features with supervised learning algorithms.
These features can be seen as a more global representation, compared to locally derived features.


In Natural Language Processing, there are new algorithms developed based on neural networks that are capable of learning a representation of the bag-of-word features into a continuous bag-of-words representation~\citep{bengio2003neural}.
This continuous bag-of-words representation can place terms with similar meaning closer and typically tend to work with lower dimensionality~\citep{mikolov2013efficient}, e.g. 100 dimensions.
Furthermore, this representation is more compact compared to the sparse bag-of-words.

Word embeddings has been previously used in WSD.\citep{chen2014emnlp,rothe2015acl} show approaches using word embeddings in knowledge-based approaches obtaining state-of-the art performance.~\citep{taghipour2015naacl,sugawara2016pacling} recently experimented with several features with SVM in supervised WSD improving more traditional features. In our work, we explore word embeddings in biomedical word sense disambiguation.

Word embeddings have been used with recurrent neural networks. Some advantages of using word embeddings is the lower dimensionality compared to bag-of-words and that words close in meaning are closer in the word embedding space. Very recent work still under preprint on using a special kind of recurrent network named LSTM (Long Short Term Memory) for WSD is recently being made available~\citep{yuan2016word} and with bidirectional LSTM~\citep{kaageback2016word}, improving over more traditional supervised learning methods.

\section{Methods}

We have compared several feature types, which are explained in more detail in this section.
These feature types range from standard unigram and bigram features to more sophisticated ones based on word embeddings.

\subsection{Evaluation Data Sets}

We evaluate the different feature sets using the MSH WSD dataset~\citep{jimeno2011exploiting} \edits{and the NLM WSD data set~\cite{weeber2001developing}.
Both data sets are available from https://wsd.nlm.nih.gov.}

\subsubsection{MSH WSD data set}

\edits{MSH WSD was automatically developed by first screening the UMLS Metathesaurus to identify ambiguous terms whose possible senses consist of two or more MeSH headings. Each ambiguous term and its corresponding MeSH heading is used to extract MEDLINE citations where the term and only one of the MeSH headings co-occur, based on the MeSH headings assigned to the citation. The term found in the MEDLINE citation is automatically assigned the UMLS CUI from the 2009AB UMLS version linked to the MeSH heading. }

MSH WSD contains~\replace{203 ambiguous terms and acronyms}{106 ambiguous abbreviations, 88 ambiguous terms and 9 which are a combination of both, for a total of 203 ambiguous entities.}~\edits{For each one of these entities, the data set contains a maximum of 100 instances per sense obtained} from the 2010 MEDLINE baseline.
Each target word contains approximately 187 instances, has 2.08 possible concepts and has a 54.5\% majority sense.
Previous supervised WSD results using Na\"ive Bayes showed a macro average accuracy over 93\%~\citep{jimeno2011exploiting}.

\subsubsection{NLM WSD data set}

\edits{The NLM WSD data set~\cite{weeber2001developing} has been used to conduct the experiments.
This set contains 50 ambiguous terms that have been manually annotated with a sense number.
Each sense number has been related to UMLS semantic types, thus originally no UMLS concept identifiers were assigned to the senses.
100 manually disambiguated cases are provided for each term.
In case no UMLS concept is appropriate,~\textit{None of the above} has been assigned.}

\edits{The selection of the 50 ambiguous words was based on an ambiguity study of 409,337 citations added to the database in 1998.
MetaMap was used to annotate UMLS concepts in the titles and abstracts based on the 1999 version of the UMLS.
50 highly frequent ambiguous strings were selected for inclusion in the test collection.
Out of 4,051,445 ambiguous cases found in these citations, 552,153 cases are represented by these 50 terms.
This means that this data set focuses on highly frequent cases.
}


\edits{We have considered the same setup as~\cite{jimeno2011exploiting} and discarded the~\textit{None of the above} category.
Since the ambiguous term~\textit{association} has been assigned entirely to~\textit{None of the above}, it has been discarded.
Furthermore, there some ambiguous words in which only one of the senses was annotated, thus it is not interesting to test machine learning methods on those words.
These words are: \textit{depression}, \textit{determination}, \textit{fit}, \textit{fluid}, \textit{frequency}, \textit{pressure}, \textit{resistance} and \textit{scale}.
This means that we will present results for 41 out of the 50 ambiguous terms.
Using the maximum frequency sense for each ambiguous word, the macro accuracy is 82.63 and the micro accuracy is 82.31.}

\subsection{Text based features}

Citations text was extracted from the title and abstract fields.
Further processing was done to the text that included lower casing, tokenization using a custom regular expression and stemming using Porter stemmer.
Unigrams and bigrams were extracted from the text and experiments with bigrams were run in combination with unigrams.

Text was processed as well to add semantic annotations in addition to local features.
UMLS concept identifiers (CUIs)~\citep{mcinnes2007using} were extracted from the MEDLINE{\tiny $^{\textregistered}$} Baseline (\url{http://ii.nlm.nih.gov/MMBaseline/index.shtml}), which is available with a version annotated with the MetaMap tool~\citep{aronson2010overview}.
In this case, the context of the ambiguous word is represented by a bag-of-concepts instead of a bag-of-words.
Another representation derived from the conceptual representation is based on UMLS Semantic Types, which is obtained from the concept annotation since UMLS concepts are assigned one or more semantic type from the UMLS Semantic Network.
We have not considered meta-data since no assumption about its availability can be made.

In addition, we have considered as well two sets of features previously used in~\citep{Zhong10}.
The first one is the part-of-speech (POS), thus a syntactic feature, of the three words before and after the ambiguous words. These words need to happen in the same sentence or a null value is used. The POS has been obtained from the MedPost/SKR POS tagger available from MetaMap~\citep{smith2004medpost}.
The second one is a set of 11 local collocations features including: $C_{−2,−2}$, $C_{−1,−1}$, $C_{1,1}$,  $C_{2,2}$, $C_{−2,−1}$, $C_{−1,1}$, $C_{1,2}$, $C_{−3,−1}$, $C_{−2,1}$, $C_{−1,2}$, and $C_{1,3}$, where $C_{i,j}$ refers to an ordered sequence of words \edits{(n-grams)} in the same sentence as the ambiguous word. Offsets i and j denote the starting and ending positions of the sequence relative to the ambiguous word. A negative or positive offset refers to a word to the left or right of the ambiguous word respectively.

\edits{Table~\ref{tab:example-features} shows example features for the ambiguous word {\t nutrition} from the citation with PubMed identifier 9336574.}

\subsection{Word embeddings}

Word embedding approaches transform the bag-of-words representation typically used in Natural Language Processing to a continuous space representation.
There are some advantages to this continuous space since the dimensionality is largely reduced and words closer in meaning are close in this new continuous space.
Existing applications to generate these embeddings based on neural networks include word2vec~\citep{mikolov2013efficient} (\url{https://code.google.com/p/word2vec}) and glove (\url{http://nlp.stanford.edu/projects/glove}).

\begin{figure}[!ht]
  \centering
    \includegraphics[width=1\textwidth]{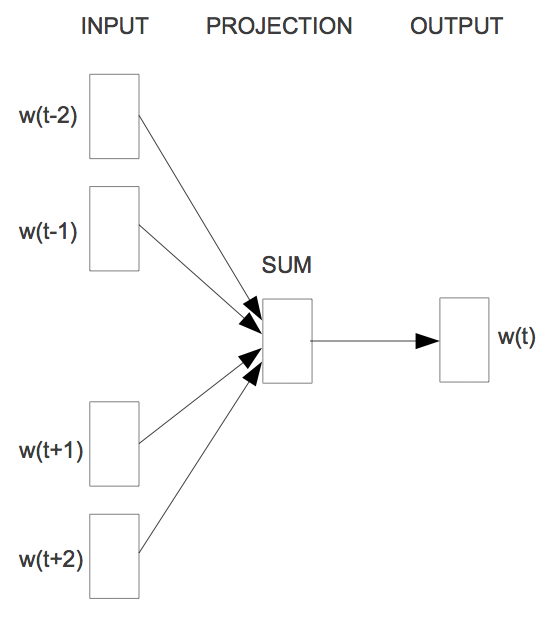}
  \caption{Continuous bag-of-words estimation diagram~\citep{mikolov2013efficient}}
  \label{fig:cbow}
\end{figure}

We have used word2vec, which offers two possible ways to generate the word embeddings. The first one is called CBOW (continuous bag-of-words). The second one is skip-gram. In this work, we have used the CBOW approach, which exemplified in Figure~\ref{fig:cbow}. In this approach, a neural network is trained to predict a word $W(t)$ given the words in the context in a supervised method.


We have experimented with CBOW word2vec vectors of several dimensions (100 to 500) and the window from which the terms are used to build the embeddings (5 to 150).

\subsubsection{Generation of word embeddings}

2014 MEDLINE is the corpus used to generate the word embeddings, which contains over 22M citations.
From this corpus, we removed the citations that appear in the disambiguation data set used in the experiments, presented later in this section.

\subsubsection{Aggregation of word embeddings}

\begin{figure}[!ht]
  \centering
    \includegraphics[width=0.5\textwidth]{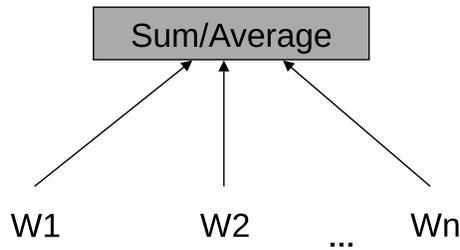}
  \caption{Aggregation of continuous bag-of-words representation vectors}
  \label{fig:vector_aggregation}
\end{figure}

After the word embeddings are generated, for each word in the dictionary a vector in an n-dimensional space is available using a look up function.
Prior to using the vectors in a machine learning method, the vectors from each individual word need to be combined.
We have evaluated the following two methods, described as well in Figure~\ref{fig:vector_aggregation}.
The whole citation text has been considered for disambiguation, thus different disambiguation context length are considered that might support the use of an average method versus one based on the sum of the vectors.

\begin{itemize}

\item Sum the vectors of the words in the context of the ambiguous word. The dimensionality of this sum is the same as the vectors generated by~\textit{word2vec}. The disambiguation context is the abstract in which the ambiguous word appears, thus it is affected by the context size.

\item Average the vectors of the words in the context of the ambiguous word. The dimensionality of the average is the same as the vectors generated by~\textit{word2vec}. This aggregation method accounts for different context sizes.

\end{itemize}

%




\edits{Table~\ref{tab:example-features} shows examples of features for the ambiguous word {\t nutrition} from the citation with PubMed identifier 9336574.}

\begin{table}[ht]
\begin{centering}
\begin{tabular}{ll}
\hline
Feature                 & Value \\
\hline
Unigrams text snippet   & Charting by exception: ... Documentation \\
                        & varies with the level of nutrition care. \\
Context                 & ... risk or with \textit{nutrition} education\\
                        & needs are ... \\

Context features        & or (-2), with (-1), education (+1), needs \\ 
                        & (+2), or (-2) with (-1), with (-1) education \\
												& (+1), education (+1) needs (+2), risk (-3) \\
												& or (-2) with (-1), or (-2) with (-1) education \\
												& (+1), with (-1) education (+1) needs (+2), \\
												& education (+1) needs (+2) are (+3) \\

POS                     & noun (-3), prep (-1), conj (-2), noun (+1), \\
                        & noun (+2), aux (+3) \\

Concepts (UMLS CUI list)& C0684240, C1554961, C1705847, C0037633, ..., \\
                        & C0525069 \\

Semantic types          & inpr, idcn, cnce, sbst, bodm, ..., orgf, orga \\

Word embedding          & -82.9220,105.5030,...,-37.6584 (150 dimensions) \\
\hline
\end{tabular}
\par\end{centering}
\caption{Example features for ambiguous word~\textit{nutrition} from citation with PubMed identifier 9336574.}
\label{tab:example-features}
\end{table}

\subsection{Supervised Learning Algorithms}

The supervised learning algorithms considered in this work are linear Support Vector Machines (SVM) based on SMO (Sequential Minimal Optimization)~\citep{platt1998sequential} using a linear kernel and feature normalization and Na\"ive Bayes (NB)~\citep{john1995estimating}, which are typically considered for this task.
We have used the implementation provided by WEKA~\citep{hall2009weka} of these algorithms for our experiments.
In addition, we have considered as well k-nearest neighbors (KNN) using cosine similarity on normalized features. 1, 3 and 5 k-nearest neighbors have been considered.

For each ambiguous word, a classifier is trained to recognize each one of the possible senses of that word.

\subsection{Long Short Term Memory}

In addition to non-deep-network learning algorithms, we have used the word embeddings to train a neural network based on a Long Short Term Memory (LSTM) unit~\citep{hochreiter1997long}.
As in the case of non-deep-network methods, one LSTM based classifier is trained per ambiguous word.
LSTM is a recurrent network that does not suffer from the~\textit{vanishing gradient}~\citep{bengio1994learning} problem and has been used in Natural Language Processing tasks~\citep{zhang2015character, sutskever2014sequence}.

LSTM units introduce mechanisms to avoid the~\textit{vanishing gradient} problem using, for a given time $t$, an input gate $i_t$, an output gate $o_t$, a forget gate $f_t$ and a cell $c_t$. The weights for these three gates and memory cell that are trained using backpropagation using training data. The input to the LSTM cell is the vector $x_t$ and the hidden output is $h_t$. The capability of LSTM of effectively dealing with long dependencies, e.g. syntactic dependencies, which might be useful to perform text analytics tasks such as disambiguation.

\begin{figure}[!ht]
  \centering
    \includegraphics[width=1.0\textwidth]{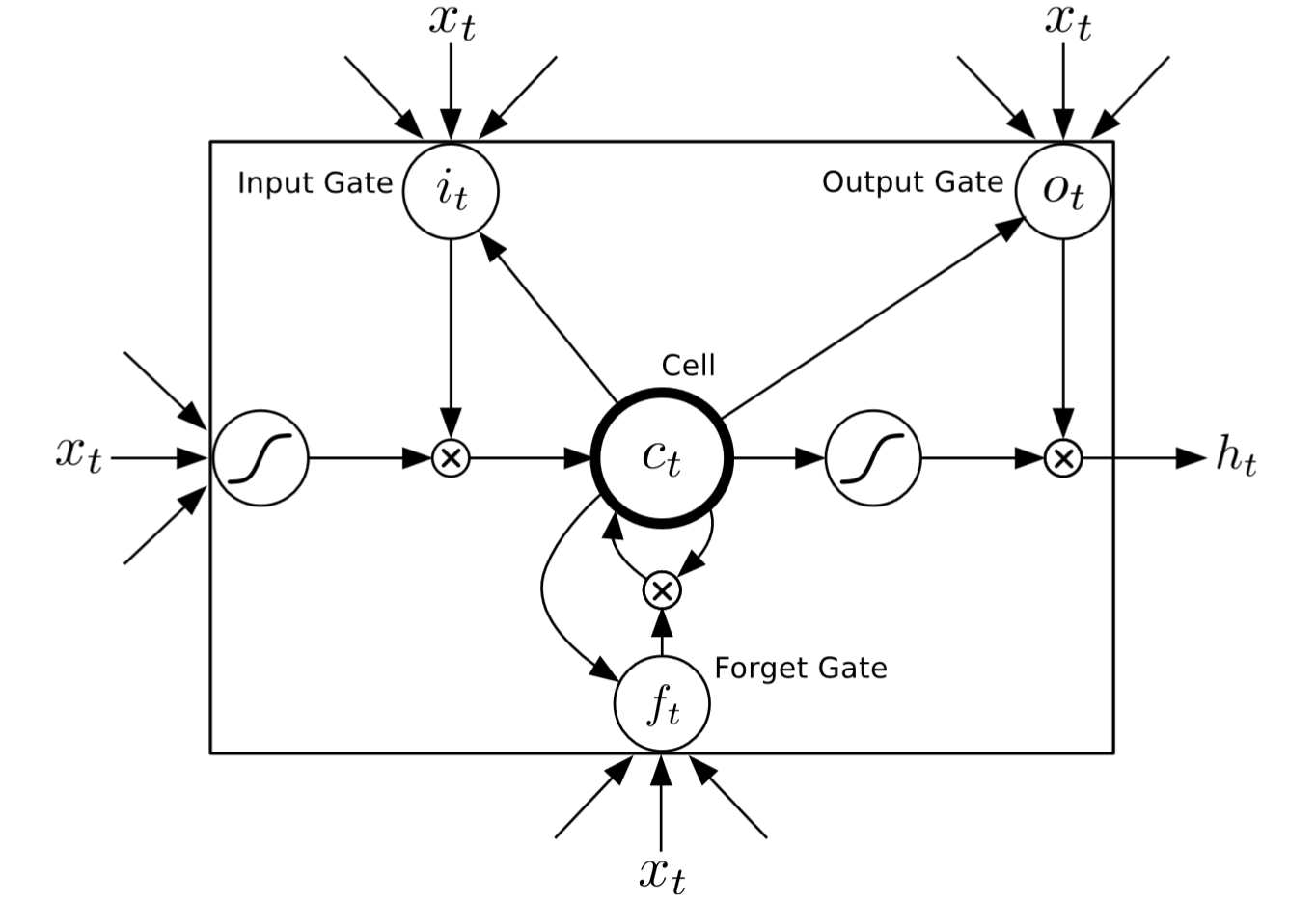}
  \caption{LSTM memory cell unit diagram~\citep{graves2013generating}}
  \label{fig:lstm}
\end{figure}

We follow the definition of LSTM unit introduced in~\citep{graves2013generating}, which follows the diagram in Figure~\ref{fig:lstm}. Equations~\ref{eq:lstm_i}~to~\ref{eq:lstm_h} show how the values in different LSTM components get calculated. Weights matrices $W$ have subscripts that indicate the components being related. For instance $W_{hi}$ is the weight matrix between the hidden output and the input gate.

\begin{eqnarray}
i_{t} = \sigma (W_{xi}x_{t} + W_{hi}h_{t-1} + W_{ci}c_{t-1} + b_{i})\label{eq:lstm_i}\\
f_{t} = \sigma (W_{xf}x_{t} + W_{hf}h_{t-1} + W_{cf}c_{t-1} + b_{f})\\
c_{t} = f_{t}c_{t-1} + i_{t} tanh(W_{xc}x_{t} + W_{hc}h_{t-1} + b_{c})\\
o_{t} = \sigma (W_{xo}x_t + W_{ho}h_{t-1} + W_{co}c_{t} + b_{o}) \\
h_{t} = o_{t} tanh (c_{t})
\label{eq:lstm_h}
\end{eqnarray}

\begin{figure}[!ht]
  \centering
    \includegraphics[width=0.5\textwidth]{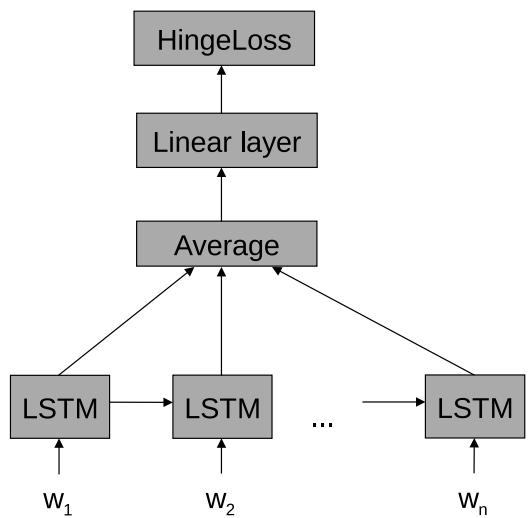}
  \caption{LSTM layout}
  \label{fig:lstm_layout}
\end{figure}

The schema of the network is shown in Figure~\ref{fig:lstm_layout} and offers a different approach to combine the word embeddings that take into account the document structure.

The size of the LSTM memory cell and the input, output and forget gates have been set as the size of the input vector defined by the word embedding size.
The output $h_t$ of the LSTM for each word in the context of the ambiguous word is averaged and a linear layer is trained to make a decision on the averaged vector, the size of the output layer is the same as the number of senses of the ambiguous word.
In the final layer, a multi-class classification Hinge loss has been used.
This network structure is similar to~\citep{zhang2015character}, which was used in text categorization, \replace{being the final layer configuration the biggest difference}{the final layer configuration being the biggest difference}.

LSTM has been implemented using Torch~\citep{collobert2011torch7} and it has been trained using AdaGrad~\citep{duchi2011adaptive}.
Learning rate has been set to 0.01 and learning rate decay to 0.01.

\edits{There are many parameters in each LSTM configuration, which may suffer from limited training data. On the other hand, there is a large quantity of unlabelled data that can be used to pre-training the recurrent neural network model. To do so, we have followed a sequence to sequence autoencoder method~\citep{dai2015semi,li2015hierarchical}. The pre-trained LSTM weights were used as initial weights of the LSTM in the supervised training instead of a random initialization. The results show a small improvement (e.g. 94.68 vs 94.64~\edits{macro accuracy} for the LSTM S100 in table~\ref{tab:word-embeddings-lstm-msh}), \replace{which}{this} \edits{difference} was not significant.}

\section{Results}

The features presented in the methods section have been evaluated using the MSH WSD, generating different feature sets.
Based on these feature sets, Na\"ive Bayes\edits{, KNN} and Support Vector Machines have been the machine learning algorithms selected to be trained for WSD.
Performance results for each feature set are presented and compared.
This section is divided in~\replace{three}{several} sections: in the first one bag-of-word features are evaluated, then the word embeddings and finally the recurrent network based on LSTM~\edits{, then several feature combination experiments are presented followed by a comparison of accuracy per ambiguous word on selected experiments}.
Selected features from each section have been combined to evaluate feature combination.

All experiments have been done using 10 fold cross-validation.
Statistical significance has been determined using a randomization version of the paired sample t-test~\citep{cohen1996empirical}.
Accuracy has been used as the evaluation measure. Macro~\edits{and micro }average have been used to aggregate the performance the ambiguous words in~\replace{the MSH WSD set}{both data sets}.
\edits{Confidence intervals 95 percent around the mean ($ \mu \pm 1.96(\frac{\sigma}{\sqrt{N}})$)are shown in selected cases in which performance of the compared methods might be close.}

\subsection{Text based features}

Table~\ref{tab:bag-of-words-msh-wsd} shows the results of training non-deep-network machine learning methods on features extracted from processing the citation text for WSD~\edits{using the MSH WSD set}.
Unigrams performance is quite competitive and \replace{only}{} in the case of Na\"ive Bayes~\edits{and KNN}, bigrams~\edits{performance differences} significantly improve the performance of unigrams.
Features such as concepts or semantic types have lower performance compared to unigrams and bigrams.
Combining the different features improves the performance of unigrams and bigrams.

In previous work \edits{on the MeSH WSD dataset}, NB has been the only machine learning method used~\citep{jimeno2011exploiting}.
Results with SVM show that the machine learning method used affects as well the accuracy with the same feature set.
KNN did not perform as well compared to the other methods.

\begin{table}[ht]
\begin{centering}
\begin{tabular}{lrrrrr}
\hline
Features/ML         &NB         &SVM        &KNN1       &KNN3       &KNN5       \\
\hline
Unigrams            &93.07/92.85&93.90/93.77&90.81/90.38&92.14/91.79&92.40/92.07\\
Bigrams             &93.87/93.76&93.94/93.86&91.26/90.90&92.75/92.41&92.91/92.57\\
POS                 &62.16/60.89&62.73/61.18&49.03/48.80&48.99/48.76&49.41/49.17\\
Collocations        &74.59/73.40&77.34/76.24&51.36/51.37&51.36/51.37&52.10/52.13\\
Concepts            &91.58/91.19&91.18/90.93&87.00/86.46&88.57/88.06&88.91/88.53\\
Semantic Types      &85.89/85.27&84.82/84.01&81.16/80.30&83.30/82.50&84.02/83.29\\
\hline
All                 &92.84/92.51&93.78/93.57&90.16/89.65&91.60/91.09&91.92/91.48\\
All-Collocations-POS&93.41/93.16&{\bf{94.36}}/{\bf{94.19}}&90.97/90.52&92.24/91.81&92.45/92.05\\
\hline
\end{tabular}
\par\end{centering}
\caption{Macro~\edits{and micro} average disambiguation results~\edits{MSH WSD set}}
\label{tab:bag-of-words-msh-wsd}
\end{table}

\edits{Table~\ref{tab:bag-of-words-nlm-wsd} shows the results for the NLM WSD set. Unigrams is as well a strong feature representation, on the other hand combining different feature sets do not improve the performance over unigrams.
NB is significantly better than SVM, which is another difference compared to the MSH WSD data set.
All the results are better compared to the maximum frequency sense baseline (macro average: 82.63 and micro average: 82.31).}

\begin{table}[ht]
\begin{centering}
\begin{tabular}{lrrrrr}
\hline
Features/ML   &NB         &SVM        &KNN1       &KNN3       &KNN5 \\
\hline
Unigrams      &{\bf{88.61}}/{\bf{88.73}}&87.87/88.00&85.23/85.27&86.72/87.05&87.53/87.69 \\
Bigrams       &86.57/86.81&86.35/86.26&85.51/85.65&87.02/87.10&87.55/87.54 \\
POS           &79.06/79.09&82.72/82.66&80.80/80.77&83.12/82.98&82.95/82.75 \\
Collocations  &85.08/84.90&85.58/85.51&82.21/82.05&83.34/83.13&83.68/83.30 \\
Concepts      &88.06/88.03&86.76/86.67&85.02/85.51&86.25/86.12&86.63/86.61 \\
Semantic Types&86.90/86.76&86.12/86.15&84.05/84.08&85.44/85.48&86.12/85.97 \\
\hline
All           &87.50/87.74&87.40/87.39&86.18/85.94&87.18/87.37&87.91/87.86 \\
\hline
\end{tabular}
\par\end{centering}
\caption{\edits{Macro and micro average disambiguation results NLM WSD}}
\label{tab:bag-of-words-nlm-wsd}
\end{table}



\subsection{Word embeddings}

In the Methods section, generation of word embedding vectors was presented.
The parameters used to generate these vectors and their aggregation are used to decide the experiments to be done and are enumerated below.
Each parameter configuration has been used to train a Na\"ive Bayes\edits{, KNN} and SVM classifier.

\begin{itemize}
\item Size of vectors generated by word2vec: 100, 150, 300 and 500.
\item Window defining how many context words are being used values are: 5, 50 and 150.
\item Aggregation method: either~\textit{sum} of the vectors or their~\textit{average} is used.
\end{itemize}

Results for the different aggregations are presented in tables~\ref{tab:word-embeddings-sum}~\edits{and~\ref{tab:word-embeddings-sum-nlm} for sum} and~\edits{tables~\ref{tab:word-embeddings-ave} and~\ref{tab:word-embeddings-ave-nlm} for average}.
Averaging seems to provide better performance, with SVM obtaining better performance compared to previously published results on the MSH WSD set.
Large vector size and large window seem to boost accuracy~\edits{for the MSH WSD set while smaller vector size and mid window size seems to perform better for the NLM WSD set}.

\begin{table}[ht]
\begin{centering}
\begin{tabular}{lrrr}
\hline
NB     &W5    &W50   &W150\\
\hline
S100   &86.46/86.01&87.64/87.28&87.62/87.24\\
S150   &86.78/86.31&87.73/87.33&87.38/86.96\\
S300   &86.83/86.40&87.61/87.22&87.49/87.08\\
S500   &86.35/85.91&87.30/86.90&75.08/74.35\\
\hline
SVM    &W5    &W50   &W150\\
\hline
S100   &92.46/92.23&92.84/92.60&93.13/92.92 \\
S150   &92.85/92.59&93.11/92.88&93.18/92.98 \\
S300   &93.00/92.75&93.27/93.07&93.35/93.17 \\
S500   &93.15/92.87&{\bf{93.41}}/93.18&93.05/92.78 \\
\hline
KNN1   &W5    &W50   &W150   \\
\hline
S100   &89.74/89.25&90.39/89.94&90.55/90.11 \\
S150   &89.83/90.08&90.43/90.62&90.64/90.79 \\
S300   &89.89/90.10&90.43/90.60&90.51/90.69 \\
S500   &89.91/89.46&90.27/89.86&88.15/87.57 \\
\hline
KNN3   &W5    &W50   &W150   \\
\hline
S100   &89.91/90.08&90.74/90.89&90.81/90.96 \\
S150   &90.02/90.20&90.69/90.88&90.83/90.98 \\
S300   &90.04/90.22&90.39/90.57&90.51/90.61 \\
S500   &89.89/89.47&90.47/90.08&86.70/86.07 \\
\hline
KNN5   &W5    &W50   &W150   \\
\hline
S100   &89.82/90.06&90.51/90.72&90.57/90.75 \\
S150   &89.75/89.99&90.44/90.69&90.60/90.80 \\
S300   &89.72/89.54&90.06/90.30&90.29/90.52 \\
S500   &89.66/89.22&90.19/89.80&85.81/85.11 \\
\hline
\end{tabular}
\par\end{centering}
\caption{Macro average disambiguation results MSH WSD set. Sum of vectors for word embeddings. S indicates the vector size and W the context window used when generating the word embeddings.}
\label{tab:word-embeddings-sum}
\end{table}

\begin{table}[ht]
\begin{centering}
\begin{tabular}{lrrr}
\hline
NB     &W5    &W50   &W150\\
\hline
S100   &85.69/85.42&86.70/86.49&85.99/85.83\\
S150   &85.96/85.68&86.36/86.23&86.39/86.29\\
S300   &85.71/85.39&86.47/86.20&86.23/86.00\\
S500   &86.06/85.80&86.36/86.12&83.86/83.50\\
\hline
SVM    &W5    &W50   &W150\\
\hline
S100   &89.18/89.22&{\bf 89.73}/{\bf{89.72}}&89.18/89.31\\
S150   &89.19/89.37&89.50/89.81&89.38/89.81 \\
S300   &88.98/88.99&88.61/88.99&88.76/89.17 \\
S500   &88.91/88.99&88.33/88.70&88.08/88.15\\
\hline
KNN1   &W5    &W50   &W150   \\
\hline
S100   &86.29/89.20&86.60/86.41&86.09/85.94 \\
S150   &86.42/86.38&86.54/86.41&86.54/86.41 \\
S300   &86.67/86.47&86.38/86.55&86.40/86.61 \\
S500   &86.79/86.73&86.66/86.78&82.41/86.17 \\
\hline
KNN3   &W5    &W50   &W150   \\
\hline
S100   &86.37/86.67&87.29/87.31&87.00/86.99 \\
S150   &86.76/86.96&87.30/87.19&86.71/86.64 \\
S300   &86.98/87.10&87.29/87.22&86.85/86.76 \\
S500   &87.32/87.45&87.19/87.10&82.78/82.37 \\
\hline
KNN5   &W5    &W50   &W150   \\
\hline
S100   &87.55/87.63&87.14/87.13&86.82/86.84 \\
S150   &86.92/86.84&86.93/86.78&86.80/86.67 \\
S300   &87.14/87.02&86.46/86.38&86.75/86.61 \\
S500   &86.72/86.67&87.13/87.05&83.40/83.07 \\
\hline
\end{tabular}
\par\end{centering}
\caption{Macro average disambiguation results NLM WSD set. Sum of vectors for word embeddings. S indicates the vector size and W the context window used when generating the word embeddings.}
\label{tab:word-embeddings-sum-nlm}
\end{table}

\begin{table}[ht]
\begin{centering}
\begin{tabular}{lrrr}
\hline
NB  &W5    &W50   &W150   \\
\hline
S100&89.87/89.34&91.31/90.85&91.52/91.11 \\
S150&90.02/89.48&91.22/90.74&91.39/90.92 \\
S300&90.10/89.56&91.22/90.75&91.46/90.99 \\
S500&89.90/89.32&90.85/90.34&74.68/73.50 \\
\hline
SVM &W5    &W50   &W150   \\
\hline
S100&93.98/93.71&94.30/94.04&94.50/94.23 \\
S150&94.20/93.93&94.40/94.15&94.60/94.33 \\
S300&94.21/93.97&94.59/94.34&{\bf{94.64}}/94.42 \\
S500&94.20/93.96&{\bf{94.64}}/94.38&94.41/94.15 \\
\hline
KNN1&W5    &W50   &W150   \\
\hline
S100&90.80/91.03&91.74/91.95&91.97/92.16 \\
S150&90.98/91.20&91.80/92.01&92.13/92.32 \\
S300&91.12/91.32&91.84/92.04&92.09/92.24 \\
S500&91.16/90.67&91.87/91.42&90.11/89.59 \\
\hline
KNN3&W5    &W50   &W150   \\
\hline
S100&91.39/91.60&92.43/92.59&92.61/92.79 \\
S150&91.73/91.90&92.41/92.57&92.68/92.85 \\
S300&91.86/92.04&92.52/92.69&92.67/92.81 \\
S500&91.88/91.44&92.60/92.23&90.00/89.48 \\
\hline
KNN5&W5    &W50   &W150   \\
\hline
S100&91.44/91.62&92.47/92.65&92.60/92.78 \\
S150&91.71/91.89&92.55/92.69&92.64/92.80 \\
S300&91.76/91.95&92.47/92.67&92.65/92.80 \\
S500&91.77/91.35&92.52/92.15&89.16/88.63 \\
\hline
\end{tabular}
\par\end{centering}
\caption{Macro average disambiguation results MSH WSD set. Average of vectors for word embeddings. S indicates the vector size and W the context window used when generating the word embeddings.}
\label{tab:word-embeddings-ave}
\end{table}

\begin{table}[ht]
\begin{centering}
\begin{tabular}{lrrr}
\hline
NB  &W5    &W50   &W150   \\
\hline
S100&88.22/88.18&88.91/89.02&88.71/88.79 \\
S150&88.38/88.38&88.74/88.82&88.91/88.99\\
S300&88.51/88.56&88.67/88.73&89.06/89.17 \\
S500&88.29/88.30&88.74/88.79&85.23/85.19 \\
\hline
SVM &W5    &W50   &W150   \\
\hline
S100&90.44/90.12&90.22/90.12&89.90/89.75 \\
S150&89.83/89.60&{\bf 90.58}/{\bf 90.42}&90.27/90.12 \\
S300&89.14/88.96&89.83/89.54&90.26/90.07 \\
S500&90.02/89.92&89.76/89.63&89.56/89.57\\
\hline
KNN1&W5    &W50   &W150   \\
\hline
S100&87.44/86.93&88.06/87.66&87.64/87.28 \\
S150&87.55/87.08&88.37/87.86&88.03/87.57 \\
S300&87.71/87.19&88.12/87.66&88.19/87.74 \\
S500&87.70/87.19&88.51/88.00&85.02/84.93 \\
\hline
KNN3&W5    &W50   &W150   \\
\hline
S100&88.11/88.15&88.88/88.88&88.36/88.50 \\
S150&88.14/88.24&88.93/88.91&88.92/89.08 \\
S300&88.46/88.53&88.71/88.64&88.65/88.79 \\
S500&88.19/88.35&88.51/88.70&85.36/85.74 \\
\hline
KNN5&W5    &W50   &W150  \\
\hline
S100&88.41/88.44&88.66/88.79&88.55/88.67 \\
S150&88.32/88.35&88.41/88.50&88.53/88.61 \\
S300&88.32/88.38&88.22/88.27&88.99/89.11 \\
S500&88.27/88.32&88.46/88.53&85.23/85.16 \\
\hline
\end{tabular}
\par\end{centering}
\caption{Macro average disambiguation results NLM WSD set. Average of vectors for word embeddings. S indicates the vector size and W the context window used when generating the word embeddings.}
\label{tab:word-embeddings-ave-nlm}
\end{table}

\subsection{Recurrent network}

The LSTM network has been evaluated using vector size 100 and 500 with window 50 in the word embedding generation~\edits{ for the MSH WSD set and for vector size 150 and 500 with window 50 for the NLM WSD set}.
10-fold cross validation has been used to obtain the results for each one of the terms.

Table~\ref{tab:word-embeddings-lstm-msh} shows the result for the two set of vectors~\edits{on the MSH WSD set}.
The 500 vector size has the best performance.

\begin{table}[ht]
\begin{centering}
\begin{tabular}{lrr}
\hline
Configuration           & Macro Accuracy & Micro Accuracy\\
\hline
SVM Unigrams            & 93.90          & 93.94 \\
SVM Bigrams             & 93.94          & 93.81 \\
SVM All-Collocations-POS& 94.36          & 94.19 \\
SVM WE S100~\edits{W150}& 94.50          & 94.31 \\
LSTM S100~\edits{W150}  & 94.64          & 94.58 \\
SVM WE S500~\edits{W50 / S300 W150} & \replace{94.52}{94.64}& 94.49 \\
LSTM S500~\edits{W50}   &{\bf{94.87}}    &{\bf{94.78}}\\
\hline
\end{tabular}
\par\end{centering}
\caption{Macro and micro average LSTM results compared to SVM unigram and bigrams and word embeddings MSH WSD set.}
\label{tab:word-embeddings-lstm-msh}
\end{table}

\edits{Table~\ref{tab:word-embeddings-lstm-nlm} shows the result for the two set of vectors on the NLM WSD set.
Using word embeddings significantly improve over using unigrams.
LSTM shows some improvement in macro averaging. which is not significant compared to SVM and word embeddings.
}

\begin{table}[ht]
\begin{centering}
\begin{tabular}{lrr}
\hline
Configuration           & Macro Accuracy & Micro Accuracy\\
\hline
SVM Unigrams            & 87.87 & 88.00 \\
NB Unigrams             & 88.61 & 88.73 \\
SVM WE S150 W50         & 90.58 & 90.42 \\
LSTM S150 W50           & 90.63 & 90.02 \\
SVM WE S500 W50         & 89.79 & 89.63 \\
LSTM S500 W50           & 90.64 & 90.19 \\
\hline
\end{tabular}
\par\end{centering}
\caption{Macro and micro average LSTM results compared to SVM unigram and bigrams and word embeddings NLM WSD set.}
\label{tab:word-embeddings-lstm-nlm}
\end{table}



\subsection{Feature combination results}

We have evaluated several feature sets in this study.
Table~\ref{tab:we-bag-of-words-msh-wsd} shows the performance of an SVM classifier in different combinations of these features with word embeddings.
As shown above word embedding features and unigrams show a significant \replace{strong}{difference in} performance.

~\edits{Since unigrams and word embeddings features have their own strengths depending on the ambiguous word, we have combined them with the expectation that the learning algorithm identifies the more relevant features for each ambiguous word during training~\citep{gabrilovich2007computing}.
The selected word embeddings used in this combination has been generated window size 50 and vector size 500 and average aggregation for the MSH WSD set and window size 50 and vector size 150 for the NLM WSD set.}

\edits{On the MSH WSD set, the accuracy obtained using SVM using this combination is 95.97/95.80.
Th\replace{is result}{e difference of results} is statistically significant ($p<0.0001$) when compared to any other evaluated method,~\edits{except for WE+ST+Concepts+Unigrams (95.95/95.80; $p<0.48$, $0.0117 \pm 0.4525$)}.
The combination of local features derived from the context of the ambiguous word and global features, provides a significant boost and sets a new performance on the MSH WSD set.
Similar behaviour has been observed by~\citep{sugawara2016pacling}.}

\edits{Other feature combinations either have a similar performance (e.g. when combining all of them as in~\textit{WE+ALL}) or show a significant lower \replace{performance}{difference in performance} when using semantic types (e.g.~\textit{WE+Semantic Types}), which had already shown lower performance when used alone.}

\begin{table}[ht]
\begin{centering}
\begin{tabular}{lrr}
\hline
Features            &Macro Accuracy & Micro Accuracy \\
\hline
WE+Unigrams         &{\bf 95.97}    &95.80\\
WE+Bigrams          &95.56          &95.40\\
WE+Concepts         &95.09          &94.92\\
WE+Semantic Types   &93.95          &93.69\\
WE+POS              &93.78          &93.50\\
WE+Collocations     &94.55          &94.33\\
\hline
WE+All              &95.00          &94.78\\
WE\edits{+ST}+Concepts+Unigrams&95.95          &95.80\\
WE+All-Collocations-POS&95.82       &95.65\\
\hline
\end{tabular}
\par\end{centering}
\caption{Macro~\edits{and micro} average feature combination study of different feature combinations including word embeddings~\edits{MSH WSD}. The learning algorithm is SVM and the word embedding configurations use 500 dimensions (S) and context window (W) of 50.}
\label{tab:we-bag-of-words-msh-wsd}
\end{table}

\edits{On the NLM WSD set, we find that the feature combination of unigrams and word embeddings and concepts and word embeddings improve slightly on using unigrams alone, which is not significant ($p<0.30$, $0.1795 \pm 0.6646$). On the other hand, the performance of the other combinations is lower compared to using unigrams alone with NB.}

\begin{table}[ht]
\begin{centering}
\begin{tabular}{lrr}
\hline
Features            &NB   &SVM  \\
\hline
WE+Unigrams         &88.79/88.91&88.69/88.85\\
WE+Bigrams          &87.50/87.54&86.86/86.84 \\
WE+Concepts         &88.73/88.76&87.64/87.57\\
WE+Semantic Types   &87.45/87.37&88.31/88.21\\
WE+POS              &86.76/86.58&88.85/88.93\\
WE+Collocations     &87.22/87.08&89.91/89.89\\
\hline
WE+All              &87.90/87.89&87.40/87.39 \\
\hline
\end{tabular}
\par\end{centering}
\caption{\edits{Macro~\edits{ and micro} average feature combination study of different feature combinations including word embeddings NLM WSD.}} 
\label{tab:we-bag-of-words-nlm-wsd}
\end{table}

\subsection{Per ambiguous word accuracy differences}

\begin{figure}[!ht]
  \centering
    \includegraphics[width=0.8\textwidth]{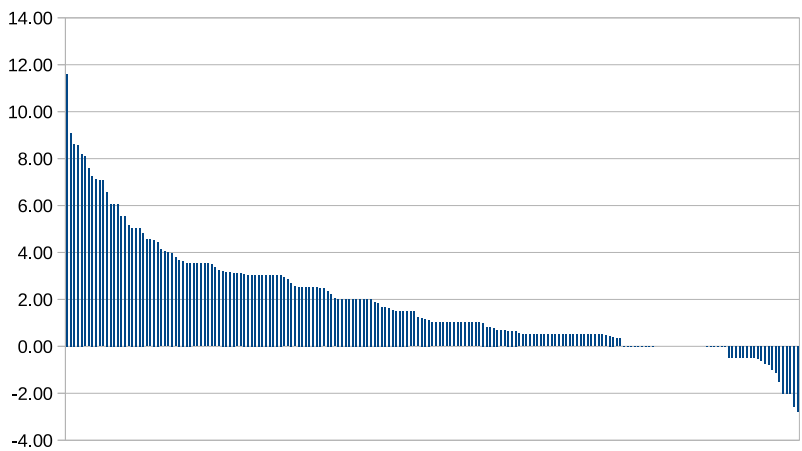}
  \caption{\replace{D}{MSH WSD set d}ifference in accuracy per ambiguous word between the combination of word embeddings with unigrams \edits{(WE+Unigrams in Table~\ref{tab:we-bag-of-words-msh-wsd})} versus just using \edits{SVM and} unigrams \edits{(Table~\ref{tab:bag-of-words-msh-wsd})} sorted in descending order.}
  \label{fig:comb-diff-unigram}
\end{figure}

\begin{figure}[!ht]
  \centering
    \includegraphics[width=0.8\textwidth]{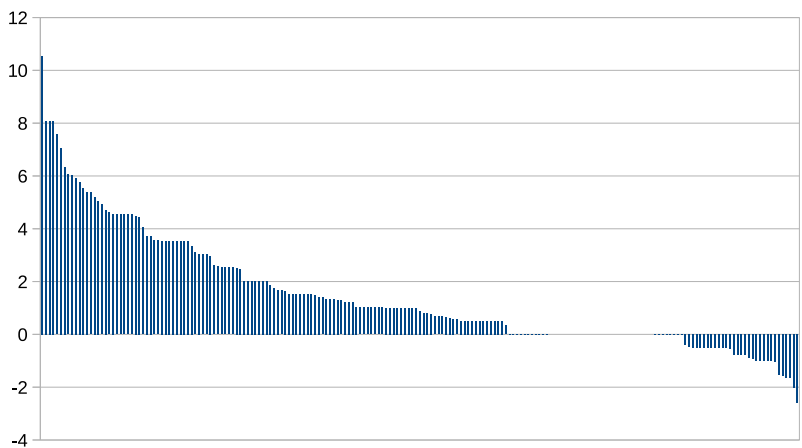}
  \caption{\replace{D}{MSH WSD set d}ifference in accuracy per ambiguous word between the combination of word embeddings with unigrams \edits{(WE+Unigrams in Table~\ref{tab:we-bag-of-words-msh-wsd})} versus average word embeddings with vectors size 500 and window 50 and SVM \edits{(Table~\ref{tab:word-embeddings-ave})} sorted in descending order.}
  \label{fig:comb-diff-svm}
\end{figure}

\begin{figure}[!ht]
  \centering
    \includegraphics[width=0.8\textwidth]{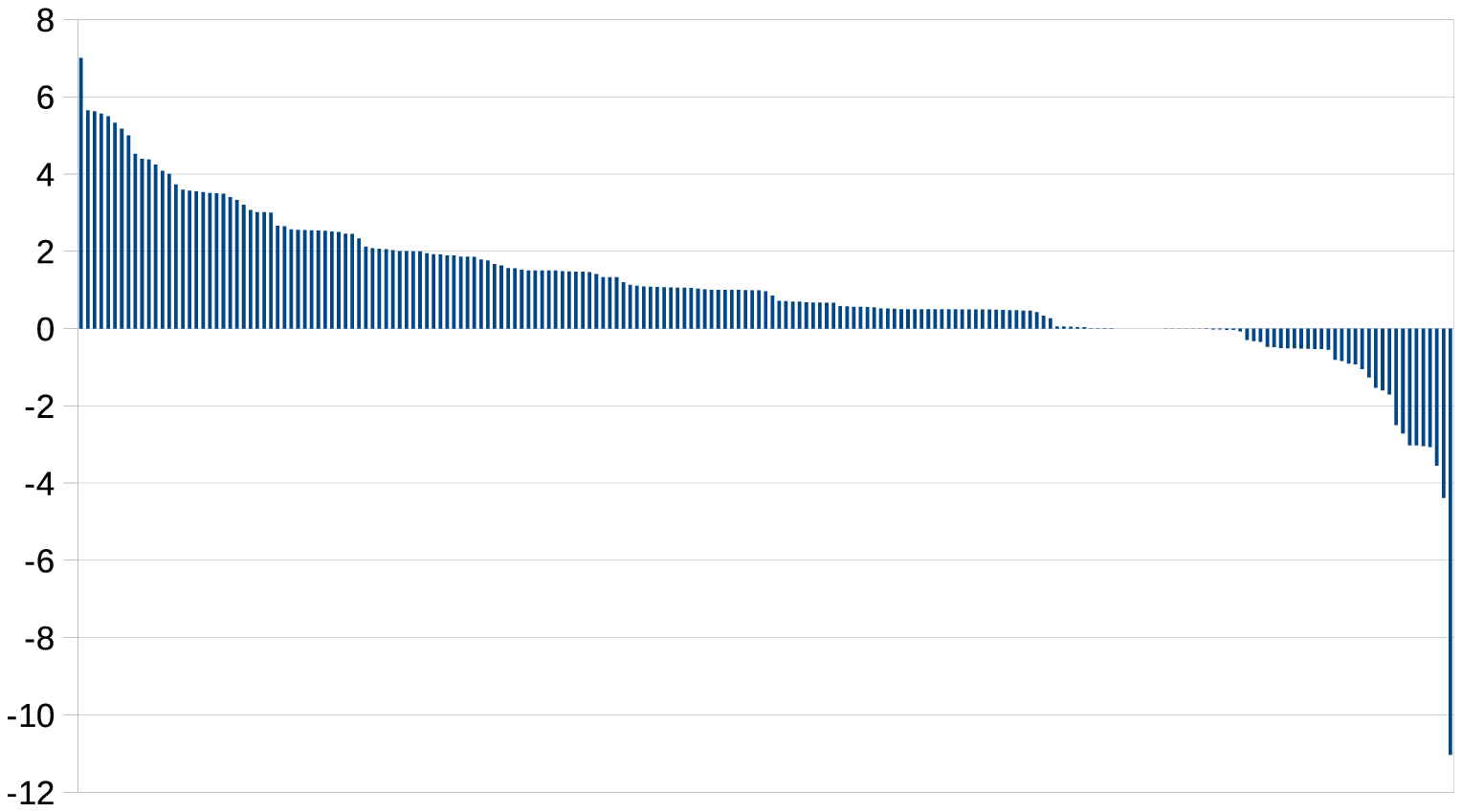}
  \caption{\replace{D}{MSH WSD set d}ifference in accuracy per ambiguous word between the combination of word embeddings with unigrams \edits{(WE+Unigrams in Table~\ref{tab:we-bag-of-words-msh-wsd})} versus LSTM with vector size 500 and window 50 \edits{(LSTM 500 in Table~\ref{tab:word-embeddings-lstm-msh})} sorted in descending order.}
  \label{fig:comb-diff-lstm}
\end{figure}

\edits{We have further examined the difference in performance for the MSH WSD feature sets and LSTM.}
Figures~\ref{fig:comb-diff-unigram},~\ref{fig:comb-diff-svm} and~\ref{fig:comb-diff-lstm} show the difference in accuracy per ambiguous term considered in this work.
In most cases, the outcome of the combination improves the results obtained by either using unigrams and SVM (Figure~\ref{fig:comb-diff-unigram}), average word embeddings with vectors size 500 and window 50 (Figure~\ref{fig:comb-diff-svm}) and LSTM 500 with vector size 500 and window 50 (Figure~\ref{fig:comb-diff-lstm}).
The differences in favour of the combination are more prominent when compared to unigram results with terms like~\textit{nursing} and~\textit{yellow fever} with the largest differences.
Compared to word embeddings, the combination performs better in most cases.
Despite the combination performing better compared to LSTM, LSTM outperforms largely the combination in ambiguous words such as~\textit{borrelia},~\textit{cement} or~\textit{WT1}.

\begin{figure}[!ht]
  \centering
    \includegraphics[width=0.8\textwidth]{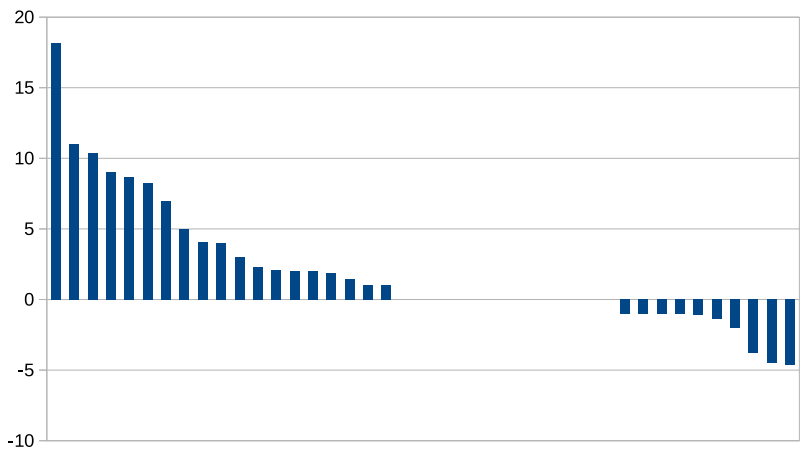}
  \caption{NLM WSD set difference in accuracy per ambiguous word between the word embeddings with SVM (SVM WE S150 W50 in Table~\ref{tab:word-embeddings-lstm-nlm}) versus unigrams and Na\"ive Bayes (NB Unigrams in Table~\ref{tab:word-embeddings-lstm-nlm}) sorted in descending order.}
  \label{fig:comb-diff-nlm-we-unigram}
\end{figure}

\begin{figure}[!ht]
  \centering
    \includegraphics[width=0.8\textwidth]{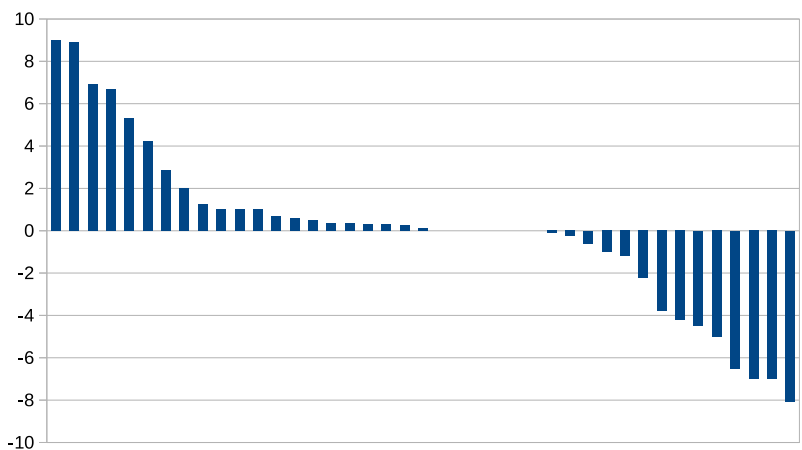}
  \caption{NLM WSD set difference in accuracy per ambiguous word between LSTM (LSTM S150 W50 in Table~\ref{tab:we-bag-of-words-msh-wsd}) versus word embeddings with SVM (SVM WE S150 W50 in Table~\ref{tab:word-embeddings-lstm-msh}) sorted in descending order.}
  \label{fig:comb-diff-nlm-lstm-we}
\end{figure}

\edits{The same analysis was done on the NLM WSD set. Feature combination does not seem to improve compared to unigrams, even when combined with word embeddings. Figure~\ref{fig:comb-diff-nlm-we-unigram} shows the difference in accuracy between SVM with word embeddings and Na\"ive Bayes and unigrams. Ambiguous word~\textit{reduction} has over 18 points difference, this has two senses\footnote{https://wsd.nlm.nih.gov/info/wsd.cases\_Final.pdf}, one as~\textit{Natural phenomenon or process} and another one as~\textit{Health Care Activity}. Word embeddings might provide means to understand the context of the ambiguous word as either related to one sense or the other. Figure~\ref{fig:comb-diff-nlm-lstm-we} shows the differences in performance bet week LSTM WE S150 W50 and SVM with word embeddings. Differences are not as large as in the previous figure. With respect to the ambiguous word~\textit{reduction}, both methods have the same performance.}

\section{Discussion}

We show that features types investigated in our work derived from MEDLINE, using word embeddings \replace{or second order features}{}, in combination with non-deep-network machine learning algorithms improve results obtained with unigrams\replace{ or 1st order features}{}.

Averaging of word embeddings with SVM improves WSD performance compared to more traditional features.
The improvement~\edits{difference} is statistically significant ($p<0.03$) when the window size is larger~\edits{or equal} than 50 and vector size is lager than 100.
Summing word embedding vectors seems to decrease performance and might be due to the effect of longer citations, the disambiguation context in this work is defined as the whole citation text.
\edits{KNN results sit in between SVM and Na\"ive Bayes and performance using word embeddings for number of neighbors $> 1$ is just above using unigrams~\edits{for the MSH WSD set}.}
Na\"ive Bayes performance is below state of the art results using word embeddings.
Weka uses the method presented in~\citep{john1995estimating} for Na\"ive Bayes, which assumes that numerical attributes are generated by a Gaussian distribution.
Loss in performance might indicate that the attributes in this new space do not follow a Gaussian distribution.

As shown in tables~\ref{tab:word-embeddings-sum} and~\ref{tab:word-embeddings-ave}, typically a larger window and vector size will improve WSD performance when using non-deep-network learning methods.


Representations such as word embeddings allow using more complex learning algorithms, such as recurrent networks.
In our work, we have shown that LSTM improves the performance of non-deep-network learning algorithms using word embeddings.
\edits{In the MSH WSD set, } the \replace{increase is}{difference in} performance is statistically significant with respect to other methods ($p<0.005$;~\edits{ci with SVM S500 W50: $0.4024 \pm 0.3130$}), even though the \replace{performance}{differences} of the \replace{two}{best} LSTM configurations is not statistically significant ($p<0.17$; $0.12445 \pm 0.2506$).
\edits{In the NLM WSD set, LSTM improves the performance of non-deep network algorithms but the difference in performance is not significant.}
\edits{This may be because word embeddings already contain information for predicting a word given the context and might be seen as a kind of pretraining.}

We have observed that LSTM performed worst compared to other methods when a significantly smaller number of examples are provided. For instance,~\textit{PAC} has only 46 and 16 examples of each one of the two senses and in the case of~\textit{hemlock}, the number of examples is 57 and 20 respectively.
LSTM needs to learn a larger number of parameters, around 81,002 with word embedding vector size 100 and 2,005,002 with vector size 500.
If enough examples are provided, LSTM could potentially improve other methods.

Word embedding based methods seem to improve state of the art methods when word embedding allow a better distinction of senses, as in~\textit{nursing} (profession versus breast feeding) and~\textit{labor} (childbirth versus work).	
On the other hand, words like~\textit{Ca}\replace{ and},~\textit{digestive}~\edits{or~\textit{blood preassure}}, in which the meanings are close, word embedding performs below state of the art methods.
In these cases, a word seems to be the discriminative clue to a proper disambiguation.

\edits{We have grouped ambiguous terms according to several criteria.
Table~\ref{tab:result-ambiguity-msh} shows the performance by groups of ambiguous words with a defined group of senses.
Results are shown for words with 2 senses (189 words) and 3 senses (12 words).
Ambiguous words with 4 and 5 senses appear only one in the data set.
Macro average shows that words with 2 senses are easier to disambiguate and that words with 3 senses are slightly more complicated.
Methods relying solely on word embeddings as features seem to have a larger drop in performance between 2 word senses and 3 word senses.}


\begin{table}[ht]
\begin{centering}
\begin{tabular}{lcccc}
\hline
Method                     &2 senses & 3 senses \\
\hline
SVM Unigrams               & 94.13   & 93.59    \\
SVM WE S500 W50            & 94.75   & 93.58    \\
LSTM WE S500 W50           & 95.00   & 94.06    \\ 
SVM WE S500 W50+Unigrams   & 96.03   & 95.22    \\
\hline
\end{tabular}
\par\end{centering}
\caption{Macro average results for ambiguous words grouped by number of senses for the MSH WSD set.}
\label{tab:result-ambiguity-msh}
\end{table}

\edits{The NLM WSD data set has 34 ambiguous words annotated with 2 senses and 6 with 3 senses.
The word~\textit{cold} is the only word with 4 senses annotated.
Table~\ref{tab:result-ambiguity-nlm} shows the results of ambiguous words grouped by number of senses.
We find that words with 2 senses are typically disambiguated with higher accuracy, while 3 senses seem to be disambiguated with lower accuracy.
There is a top of 100 examples for each ambiguous word, so if 3 senses appear, there is less training data per ambiguous word sense.
}

\begin{table}[ht]
\begin{centering}
\begin{tabular}{lcc}
\hline
Method                     &2 senses &3 senses \\
\hline
NB unigrams                &90.87    &75.15    \\
SVM Unigrams               &90.03    &74.81    \\
SVM WE S150 W50            &91.25    &73.68    \\
LSTM WE S150 W50           &93.40    &74.37    \\ 
LSTM WE S500 W50           &93.29    &75.20    \\ 
\hline
\end{tabular}
\par\end{centering}
\caption{Macro average results for ambiguous words grouped by number of senses for the NLM WSD set.}
\label{tab:result-ambiguity-nlm}
\end{table}



\edits{As defined in~\cite{jimeno2011exploiting}, the ambiguous words in MSH WSD can be divided into terms (T), abbreviatios (A) and words that might act as both (AT).
Table~\ref{tab:result-term-msh} shows the macro average performance on these sets of words.
Terms (T) are the most difficult to disambiguate, while abbreviations seem to be the easiest set.
SVM with word embeddings and unigrams performs the best on all the categories.
LSTM seems to be better for terms compared to SVM when unigrams and word embeddings are used separately.
This different seems to be less clear for abbreviations and ATs.
SVM with word embeddings seems to perform less well for the AT group in comparison to the performance in other categories with other methods.}

\begin{table}[ht]
\begin{centering}
\begin{tabular}{lccc}
\hline
Method                     & T     & A     & AT     \\
\hline
SVM Unigrams               & 90.23 & 97.26 & 94.55 \\
SVM WE S500 W50            & 90.92 & 97.58 & 93.69 \\
LSTM WE S500 W50           & 92.04 & 97.34 & 94.64 \\ 
SVM WE S500 W50+Unigrams   & 93.07 & 98.32 & 96.50 \\
\hline
\end{tabular}
\par\end{centering}
\caption{Macro average by ambiguous word type (Term (T), Abbreviation (A), Term-Abbreviation (TA)) for the MSH WSD set.}
\label{tab:result-term-msh}
\end{table}

\section{Conclusions and Future Work}




The combination of unigrams \replace{(local features)}{} and word embeddings \replace{(global features)}{} sets a new state of the art performance with the MSH WSD data set with an accuracy of 95.97~\edits{, but this is not the case for the NLM WSD set}.
\edits{For the NLM WSD set, LSTM with word embeddings provides the better accuracy followed by non-deep-network learning algorithms with word embeddings and feature combination does not seem to improve performance.
On both sets, word embeddings and LSTM improve over single feature sets.}

Using representations based on word embeddings reduce the dimensionality of the bag-of-word vectors and could be used in functions for probability estimation, which could be used in unsupervised methods based on probabilistic graphical models~\citep{jimeno2014knowledge}.

Recent work has studied the use of not only generation of vectors at the word level but at the document level, for instance for text categorization~\citep{le2014distributed,kosmopoulos2015biomedical} and it would be interesting to see the performance of their methods on the WSD problem presented in this work.

LSTM has been trained using a reduced number of examples and could benefit from using a larger set.
Training has been done on examples from the MSH WSD~\edits{and NLM WSD} data set\edits{s}.
Following the procedure used to generate \replace{this}{the MSH WSD} data set, it would be possible to extend the training set.

Supervised methods perform typically better compared to knowledge-based approaches but require training data, which limits its usability. The outcome of this work is relevant to understand how word embeddings support biomedical word sense disambiguation and encourages extending the current work in the knowledge-based scenario.

\section{Acknowledgements}

The author would like to thank Dr. Bridget McInnes for motivating this work and providing ideas and semantic annotation for the examples. 


\bibliographystyle{elsarticle-harv}
\bibliography{bibliography}  

\end{document}